\documentclass[journal]{IEEEtran}

\usepackage{amsmath,amssymb}
\usepackage{booktabs}
\usepackage{array}
\usepackage{graphicx}
\usepackage{url}
\usepackage[hidelinks]{hyperref}

\begin{document}

\title{Register Bias in Complexity-Based Large Language Model Routing}
\author{Simran Koul
\thanks{S. Koul is an independent researcher (email: simrankoul2026@gmail.com).}}

\markboth{}{Koul: Register Bias in Complexity-Based LLM Routing}
\maketitle

\begin{abstract}
Large language model services increasingly route each query to one of several models of
differing capability, using a cheap estimate of query complexity to send easy queries to
small models and hard queries to large ones. I show that this routing step is not register
neutral: text written in a non-standard English register, African American English or the
English of second-language writers, is systematically assigned a lower-capacity tier than a
meaning-equivalent standard-English version of the same query. The effect is driven by a
specific, common routing signal, input length, because non-standard registers omit function
words and thus look shorter and therefore simpler; other complexity signals do not carry it.
I demonstrate the disparity on 37{,}704 authentic learner sentence pairs and on a controlled
parallel corpus. I then measure the quality consequence on a device, edge, and cloud model
ladder and find that the harm is driven by pervasive model bias, every tier, including a
frontier cloud model, answers non-standard-register queries significantly less accurately,
while the marginal quality cost of the routing decision itself is not significant on this
benchmark. Complexity-based routing thus compounds the exposure of the users that the models
already serve worst.
\end{abstract}


\section{Introduction}
\IEEEPARstart{C}{ost-aware} serving of large language models (LLMs) routes each query to a
model of appropriate strength, sending easy queries to small models and hard queries to large
ones based on an estimate of query difficulty~\cite{frugalgpt,routellm,faircarbon}. The
routing signal is typically cheap and surface-level. I ask whether that routing decision is
fair across the linguistic register in which a query is written, and I study the router as an
object of audit independent of any single deployed system.

I find that a complexity-based router treats meaning-equivalent queries differently by
register. A question written in African American English or in the English of a second-language
writer is assigned a lower-capacity tier than a standard-English version of the same question.
This paper makes the following contributions.
\begin{enumerate}
\item The first audit, to my knowledge, showing that a complexity-based LLM router assigns
meaning-equivalent queries to different capability tiers depending on linguistic register,
evidenced on authentic human text (37{,}704 learner pairs) and a controlled parallel corpus.
\item A decomposition of the disparity by complexity signal: token length carries it robustly,
while readability and syntactic-depth signals do not, and can even reverse.
\item A mechanistic account: non-standard registers omit function words, shortening the text,
so length-based routing reads them as simpler.
\item A quality decomposition on a device, edge, and cloud ladder separating model bias from
routing-induced harm, with the finding that even a frontier cloud model is significantly
register-biased, while the routing decision's own marginal quality cost is not significant on
this benchmark, reported honestly as a null.
\end{enumerate}

\begin{figure}[t]
\centering
\includegraphics[width=\columnwidth]{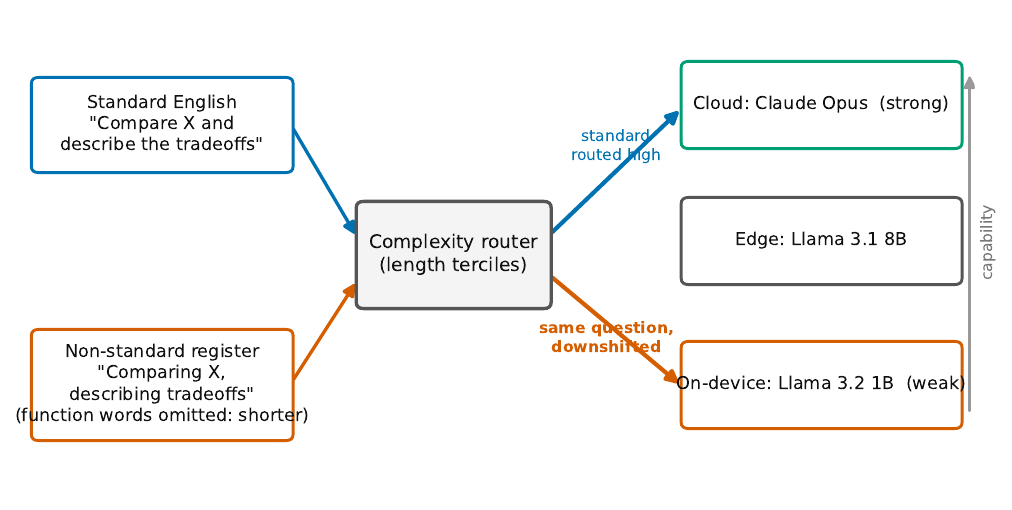}
\caption{The routing pipeline and the register-downshift mechanism. A length-based complexity
router assigns each query to a device, edge, or cloud tier. A non-standard-register phrasing of
the same question omits function words and is therefore shorter, so the router assigns it a
weaker tier: the disparity this paper audits.}
\label{fig:arch}
\end{figure}

\section{Related Work}
\textbf{LLM routing and cascades.} Cost-aware routing and model cascades reduce serving cost by
sending each query to a model of appropriate strength~\cite{frugalgpt,routellm}. Closest to this
work, a carbon-aware and fairness-aware router~\cite{faircarbon} argues that naive
energy-saving routing can leave users in certain regions or languages with lower-quality
service, and adds a distributional-fairness constraint. That work operates over geography and
distinct languages and proposes a mitigation; it does not audit register variation within
English, and it does not decompose routing-induced disparity from the underlying models' own
bias, which are the questions I take up.

\textbf{Dialect bias in LLMs.} A single model's accuracy and behavior can vary sharply by
dialect. Covert dialect prejudice has been demonstrated in modern
LLMs~\cite{hofmann2024}, and dialect stress-testing frameworks quantify accuracy gaps across
English varieties~\cite{value2022,multivalue2023}. I take single-model dialect bias as
established and do not claim it; my contribution concerns the \emph{routing} layer that sits
above the models.

\textbf{Complexity measures.} I use standard, published readability and complexity signals
(token length, Flesch-Kincaid grade~\cite{fleschkincaid}, Gunning fog~\cite{gunning}, and
dependency-parse depth) as representative of the difficulty proxies real routers use, so the
finding concerns the class of complexity-based routers rather than one bespoke formula.

\section{Method}

\subsection{Framing: complexity-based routing}
I treat the router as a function that maps a query's measured complexity to one of three
capability tiers (Fig.~\ref{fig:arch}), and I ask whether that function assigns meaning-
equivalent queries to different tiers depending on register. I audit four standard complexity signals: token
\emph{length}; maximum dependency-parse depth (\emph{syn\_depth}); Flesch-Kincaid grade
(\emph{fk\_grade}); and Gunning fog (\emph{fog}). For each signal I fit a tercile router: I set
the 33rd and 66th percentiles on a reference set of standard-English texts, then assign any
text to tier 0, 1, or 2 by which band its measure falls in. Higher measured complexity routes
to a higher tier.

\subsection{Register variation: authentic and controlled}
\textbf{Authentic (primary).} The W\&I+LOCNESS corpus~\cite{bea2019} contains real English-
learner writing in which each source sentence is a learner's original and an annotation reduces
it to a corrected, standard-English version. Reconstructing the correction yields an authentic
learner-original, standard-corrected parallel pair for the same content. I use the beginner,
intermediate, and advanced learner strata (37{,}704 pairs), with the native LOCNESS stratum as
a control.

\textbf{Controlled (secondary).} To hold the question exactly constant across registers, I take
a 300-question sample of Natural Questions~\cite{naturalquestions,nqopen} and transform each
standard-English question into African American English and Indian English variants using
Multi-VALUE~\cite{multivalue2023}, a rule-based transformer whose African American English rules
are human-validated. Questions are normalized to well-formed form before transformation;
transforms that error out are dropped (1 of 300).

\subsection{Semantic-equivalence gate}
A register variant is usable only if it asks the same question as its standard-English original.
I gate every controlled variant with an out-of-pipeline judge (Claude Opus) instructed to
compare meaning only and to ignore grammar, dialect, and spelling, keeping a question only if
every variant passes. This retained 279 of 299 questions (6.7\% dropped). The judge is
deliberately not one of the routed models.

\subsection{Routing-disparity metric}
The router's decision is deterministic given the text, so the primary result needs no model
inference. For each parallel pair I compute the tier assigned to the standard-English member and
to each register variant under each complexity measure, and count how often the non-standard
version is routed down, up, or the same. I test the asymmetry with a paired sign test (McNemar
exact for small counts; a continuity-corrected normal approximation for large counts).

\subsection{Quality tiers and grading}
For the quality analysis I use a realistic device, edge, and cloud capability ladder: Llama 3.2
1B (on-device)~\cite{llama32}, Llama 3.1 8B (edge)~\cite{llama3}, and Claude Opus (cloud
frontier)~\cite{claude}. The two open models run locally with Ollama~\cite{ollama}. I grade each
answer by fact-containment against the reference answers with light morphological normalization,
which is deterministic and register-invariant, so the quality metric itself carries no dialect
bias, and the grader is not one of the routed models.

\section{Offline Results: The Routing Disparity}

\subsection{Authentic learner text}
On 37{,}704 authentic learner/corrected pairs, the length-based router routes the learner's
original to a weaker tier than its standard correction far more often than the reverse: 1{,}272
downshifts versus 474 upshifts (paired sign test, $p<10^{-3}$). The disparity is significant and
one-directional under the length signal. The other three signals do not show it: under
syntactic depth, Flesch-Kincaid grade, and Gunning fog the counts run the other way (for
example, Flesch-Kincaid 1{,}147 down versus 2{,}194 up), because learner errors and missing
punctuation inflate readability and depth scores. On the 988 native LOCNESS pairs the length
disparity is much weaker (20 down versus 7 up, $p=0.019$) and the readability signals behave
erratically, consistent with those signals being noisy on short single sentences.

\subsection{Controlled parallel corpus}
On the 279 gated Natural-Questions parallel sets, terciles fit on the standard-English
questions, the length-based router again downshifts both non-standard registers significantly:
African American English 68 down versus 42 up ($p=0.017$), Indian English 101 down versus 34 up
($p<0.001$). The other signals are inconsistent: syntactic depth and Gunning fog are non-
significant, and Flesch-Kincaid reverses for Indian English (11 down versus 51 up, $p<0.001$).

\subsection{Mechanism}
Every downshift under the length signal has the same cause. Non-standard registers omit function
words, African American English auxiliary deletion (``when did X air'' becomes ``when X air'')
and second-language writers dropping articles and prepositions, which shortens the text. A
length-based router reads shorter as simpler and assigns a weaker tier. Two independent data
sources converge on this mechanism. The robust claim is therefore narrow and honest: token-
length-based routing, the most common routing signal, systematically misroutes non-standard-
register queries to weaker tiers, and other complexity signals do not show a consistent effect.

\section{Quality Cost of the Misrouting}
I answer each of the 279 gated register-parallel questions with the device, edge, and cloud
ladder and grade by fact-containment, decomposing the quality effect into model bias and
routing-induced harm.

\subsection{Path B: model bias (tier fixed)}
\begin{figure}[t]
\centering
\includegraphics[width=\columnwidth]{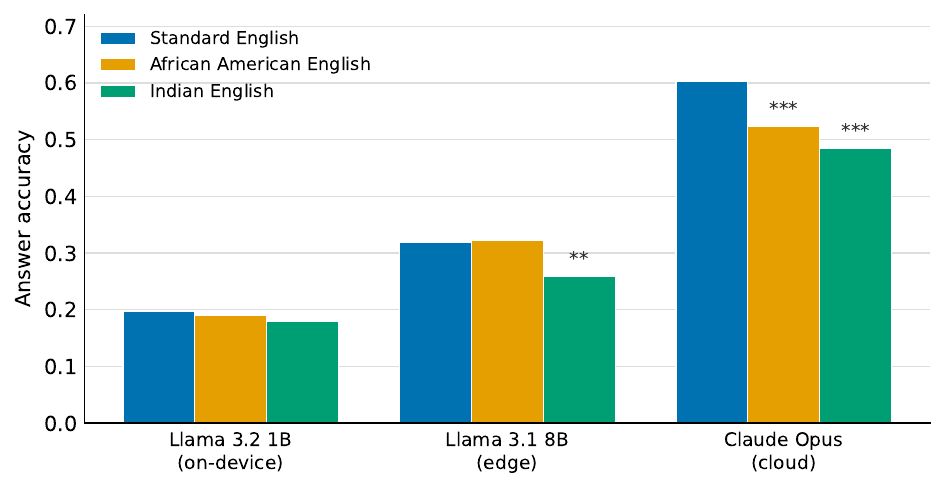}
\caption{Per-tier answer accuracy by register on the device, edge, and cloud ladder. Register
bias persists and strengthens up the ladder: even the frontier cloud model answers African
American English and Indian English significantly less accurately than Standard English. Stars:
McNemar $p$ vs.\ Standard English ($*\,p<.05$, $**\,p<.01$, $***\,p<.001$).}
\label{fig:pathb}
\end{figure}

Holding the tier fixed, I compare accuracy on the standard-English question with accuracy on
each register variant (McNemar paired test), Table~\ref{tab:pathb} and Fig.~\ref{fig:pathb}.
The models are
register-biased, and the effect strengthens with capability rather than vanishing: the frontier
cloud model, despite answering standard-English questions best, is significantly less accurate
on both African American English and Indian English. A state-of-the-art model is not
register-neutral. The 1B is too weak to show the effect (floor).

\begin{table}[t]
\caption{Path B: per-tier accuracy by register (McNemar $p$ vs.\ standard English).}
\label{tab:pathb}
\centering
\begin{tabular}{@{}lccc@{}}
\toprule
Tier & SAE & AAVE & Indian \\
\midrule
1B (on-device) & 0.197 & 0.190 ($p{=}0.87$) & 0.179 ($p{=}0.56$) \\
8B (edge)      & 0.319 & 0.323 ($p{=}1.0$)  & 0.258 ($p{=}0.010$) \\
Claude (cloud) & 0.602 & 0.523 ($p{=}0.0005$) & 0.484 ($p{<}0.001$) \\
\bottomrule
\end{tabular}
\end{table}

\subsection{Path A: routing-induced harm}
Under the length-based router, each register variant is assigned a tier by its own length. I
compare the accuracy it realizes there against the accuracy it would have received at the tier
its standard-English version routes to, on the questions the router downshifts. The effect is
not significant for either register (African American English realized 0.430 vs.\ counterfactual
0.423, $p=0.80$; Indian English realized 0.341 vs.\ counterfactual 0.348, $p=0.87$), even with a
wide capability gap between tiers. The apparent reason is the benchmark: Natural-Questions items
are short and nearly uniform in length, so the router's tier boundaries are tight and the
downshifted questions are not systematically the ones where the tier gap decides the answer. I
report the null.

\subsection{Reading the two paths together}
The quality cost borne by non-standard-register users is driven by pervasive model bias
(Path~B), which is significant and, strikingly, present even in the frontier cloud model.
Complexity-based routing compounds the exposure of these users by systematically downshifting
their queries (Section~V), sending exactly the users the models serve worst toward the weaker
tiers. The marginal quality cost of the routing decision itself is not significant on this
benchmark (Path~A), and I do not claim otherwise.

\section{On-Device Deployment Realism}
In mobile deployments the weakest tier runs on the user's own device, so the users whose
register is downshifted are the ones served by the on-device model. I verify this is a real
deployment path by cross-compiling llama.cpp~\cite{llamacpp} for Android and running Llama 3.2
1B (the same weights as the server-side tier) on a physical Samsung Galaxy S25+ over the Android
Debug Bridge, an independent path using no prior system's application. On a 40-question sample
the 1B runs on-device at a mean 29.1 tokens per second, with on-device accuracy 0.175 essentially
matching the server-side 0.200 on the same questions and correctness agreement on 37 of 40
questions (0.925). The weakest tier is therefore genuinely deployable on the phone and its
answers match the server-side model, so the quality disparity measured server-side transfers to
real hardware. This is a deployment-realism result, not a new quality measurement, since the
on-device and server tier-0 models are identical.

\section{Limitations}
Rates are workload-specific; the robust claims are the direction and the mechanism, not the exact
percentages. The controlled register variants are rule-transformed, which is disclosed;
authenticity rests on the W\&I+LOCNESS arm. Readability and syntactic-depth measures are noisy on
short single sentences, and the length result is the reliable one. The routing-induced quality
harm (Path~A) is null on a short-question benchmark; a workload with more length variation is
needed to test whether that null is benchmark-specific.

\section{Conclusion}
Complexity-based LLM routing, judged by the most common signal it uses, systematically sends
non-standard-register queries to weaker tiers. The users so downshifted are also the ones every
tier, including a frontier cloud model, answers significantly less accurately. The routing layer
thus compounds the exposure of the users the models already serve worst, a fairness concern that
sits above any individual model and that a router can be designed to avoid.

\end{document}